# Possibility and necessity functions over non-classical logics

Philippe Besnard
IRISA
Campus de Beaulieu
Avenue du Général Leclerc
35042 Rennes Cedex
France

Jérôme Lang
IRIT
Université Paul Sabatier
118 route de Narbonne
31062 Toulouse Cedex
France

## Abstract

We propose an integration of possibility theory into non-classical logics. We obtain many formal results that generalize the case where possibility and necessity functions are based on classical logic. We show how useful such an approach is by applying it to reasoning under uncertain and inconsistent information.

## 1 Introduction

Possibility theory has been widely used in Artificial Intelligence to represent uncertain knowledge in a more qualitative way than, for example, probability theory: indeed, it is equivalent to work with "quantitative" possibility theory (which means using possibility and necessity measures and possibility distributions, which map formulas or worlds to $[0,1]$) or with its qualitative counterpart (where qualitative necessity and possibility relations are preorders on the logical language and qualitative possibility distributions are just preorders on the set of worlds). Besides, its connection to various qualitative formalisms in logic and Artificial Intelligence has been established, notably with epistemic entrenchment relations in [DP 91a], conditional logics in [Bou 92] [FHL 94], System Z in [BDP 92]. The use of possibility theory in Artificial Intelligence covers non-monotonic reasoning [DP 91b], belief revision, inconsistency handling, inheritance and default rules handling, temporal reasoning, constraint satisfaction,...

In Knowledge Representation, many non-classical logics have been used (note that in this paper we consider only non-classical logics sharing the same language as classical logic). Each of them was intended for some particular focus, a specific aspect of reasoning: E.g. paraconsistent logics have been used to deal with contradictory knowledge bases. Or, intuitionistic logic has been used to take into account some subtle distinctions between statements involving double negation for example. Or, Kleene's 3-valued logic (and other many-valued logics) has been used to cope with statements for which neither truth nor falsity make sense.

Now, possibility theory brings in something more that should be fruitfully exploited as complementary to such aspects of reasoning. Hence, we study how to integrate possibility theory with non-classical logics. Our work comes from the following two facts:

- even when the involved uncertainty has a possibilistic nature, "classical" possibility theory may not be well-suited to the addressed problem, due to shortcomings, not of possibility theory itself but of classical logic, on which possibility theory is defined. For example, some problems require a formalization with a local view of inconsistency: this is impossible with classical possibility theory (we need a paraconsistent approach, cf Section 3).

- on the other hand, non-standard logics such as intuitionistic logic, paraconsistent logics,... are not expressive enough to express uncertainty in a gradual way.

These arguments show that it is generally valuable to integrate non-classical logics with a numerical theory of uncertainty. Now, the reason why we focus in this paper on possibility theory rather than another theory of uncertainty, is its qualitative nature (as it amounts to a "numerical account" of preordering relations over formulas or worlds), which should make it a priori simpler to generalize than more quantitative approaches such as probability theory or belief functions.

The methodology we follow in this paper consists of going from the general case to the particular case:

- in Section 2, we investigate whether, and under which conditions, important properties of possibility theory remain valid when generalized. We state the results in the most general case to make the study "reusable", though the applications developed in Section 3 focus on paraconsistency.

- in Section 3, we take a case study, that is, we choose a paraconsistent logic (namely $C_1$) and discuss more practical applications to reasoning with uncertain and inconsistent information.

*Research supported by CNRS in project "Gestion de l'évolutif et l'incertain dans une base de connaissances".

## 2 Non-classical necessity and possibility functions

### 2.1 Necessity/possibility functions

The natural presentation of necessity and possibility functions (see [Zad 78] for instance) shows that possibility theory consists in meta-level definitions over classical logic, which respect completely the structure of classical logic.

This suggests that similar functions could be defined on other logics than classical logic; so, replacing $(\mathcal{L}, \vdash)$ by $(\mathcal{L}, \vdash_L)$ where $L$ is a given non-classical logic, we can look for a definition of possibility/necessity functions on the logic $L$. We deal with classical propositional[1] languages, built from a list of propositional variables – sometimes required to be finite –, and the connectives $\neg, \wedge, \vee, \rightarrow, \leftrightarrow$ (where $\vdash_L \varphi \leftrightarrow \psi$ is a shorthand for $\vdash_L \varphi \rightarrow \psi$ and $\vdash_L \psi \rightarrow \varphi$). The only varying parameter is then the consequence relation $\vdash_L$. We now give a generic definition of non-classical necessity/possibility functions, of which the usual necessity/possibility functions correspond to the special case where $L$ is classical logic (Section 3 deals with the special case where $L$ is the paraconsistent logic $C_1$).

**Definition:** let $\mathcal{L}$ be a classical propositional language and $\vdash_L$ a consequence relation, $L$ being a given (maybe non-classical) logic. A L-necessity function is a mapping $N$ from $\mathcal{L}$ to $[0,1]$ satisfying the following axioms[2]

(Taut)  if $\vdash_L \varphi$ then $N(\varphi) = 1$

  (Eq)  if $\vdash_L \varphi \leftrightarrow \psi$ then $N(\varphi) = N(\psi)$

(Conj)  $N(\varphi \wedge \psi) = \min(N(\varphi), N(\psi))$

When $L$ is classical logic, we recover the classical necessity functions. Whatever the logic $L$ is, the following property always entails (Eq):

(Dom)  if $\vdash_L \varphi \rightarrow \psi$ then $N(\varphi) \leq N(\psi)$

**Proposition 1:** (Dom) is entailed by (Eq) and (Conj) on condition that $\vdash_L$ satisfies:

$$\frac{\vdash_L \varphi \rightarrow \psi}{\vdash_L (\varphi \wedge \psi) \leftrightarrow \varphi}$$

Hence, for all logics $L$ fulfilling the latter condition, a necessity function can then be characterized as a function $N : \mathcal{L} \rightarrow [0,1]$ satisfying (Taut), (Conj), (Dom).

[1] For the sake of simplicity, we consider only the propositional level throughout the paper.

[2] Many definitions of necessity functions include the extra axiom (Contr) if $\vdash \neg \varphi$ then $N(\varphi) = 0$ but not all: for example, in [DLP 94] the quantity $N() > 0$, reflects a degree of (partial) inconsistency. Note that requiring (Contr) or not –and the same for (Taut)– does not make much difference since (Dom) ensures that contradictions (resp. tautologies) have anyway the lowest (resp. highest) necessity degree. Now, the reason why we require (Taut) and not (Contr) concerns the characterization of necessity functions in terms of possibility distributions.

The dual functions of necessity functions are called possibility functions. They can be defined by 3 axioms about contradiction, equivalence and disjunction:

**Definition:** A L-possibility function is a mapping $\Pi$ from $\mathcal{L}$ to $[0,1]$ such that

(Contr)  if $\vdash_L \neg \varphi$ then $\Pi(\varphi) = 0$

  (Eq$_\Pi$)  if $\vdash_L \varphi \leftrightarrow \psi$ then $\Pi(\varphi) = \Pi(\psi)$

  (Disj)  $\Pi(\varphi \vee \psi) = \max(\Pi(\varphi), \Pi(\psi))$

Whatever the logic $L$, the next property entails (Eq$_\Pi$):

(Dom$_\Pi$)  if $\vdash_L \varphi \rightarrow \psi$ then $\Pi(\varphi) \leq \Pi(\psi)$

**Proposition 2:** (Dom$_\Pi$) is entailed by (Eq$_\Pi$) and (Disj) on condition that $\vdash_L$ satisfies:

$$\frac{\vdash_L \varphi \rightarrow \psi}{\vdash_L (\varphi \vee \psi) \leftrightarrow \psi}$$

### 2.2 Some properties of non-classical necessity/possibility functions

When $L$ is classical logic, (L-)possibility functions can be defined from (L-)necessity functions by means of $\forall \varphi \in \mathcal{L}, \Pi(\varphi) = 1 - N(\neg \varphi)$ and (L-)necessity functions can be defined from (L-)possibility functions by $\forall \varphi \in \mathcal{L}, N(\varphi) = 1 - \Pi(\neg \varphi)$. That is, "classical" necessity and possibility functions enjoy the (double) duality property:

(D1) $\Pi$ is a possibility function iff $d_\Pi : \mathcal{L} \rightarrow [0,1]$ defined by $d_\Pi(\varphi) = 1 - \Pi(\neg \varphi)$ is a necessity function.

(D2) $N$ is a necessity function iff $d_N : \mathcal{L} \rightarrow [0,1]$ defined by $d_N(\varphi) = 1 - N(\neg \varphi)$ is a possibility function.

Some questions we may ask are: how can (D1) and (D2) carry over to L-necessity and L-possibility functions? When are (D1) and (D2) equivalent?

**Proposition 3:** if $\vdash_L \neg\neg\varphi \leftrightarrow \varphi$ then (D1) $\Leftrightarrow$ (D2).

**Proposition 4:** if $L$ satisfies

1. $\vdash_L \varphi \leftrightarrow \neg\neg\varphi$
2. $\vdash_L \neg(\varphi \wedge \psi) \leftrightarrow (\neg\varphi \vee \neg\psi)$
3. $\vdash_L \neg(\varphi \vee \psi) \leftrightarrow (\neg\varphi \wedge \neg\psi)$

and the following inference rules

$$\frac{\vdash_L \varphi \quad \vdash_L \varphi \rightarrow \psi}{\vdash_L \psi} \qquad \frac{\vdash_L \varphi \rightarrow \psi}{\vdash_L \neg\psi \rightarrow \neg\varphi}$$

then (D1) and (D2) hold.

Note that among non-classical logics admitting (1)-(3) and the above two inference rules (modus ponens and contraposition), there are various relevant logics such as the logic E [AB 75]. Let us now have a look on necessary conditions for having (D1) (or (D2)).

**Proposition 5:** if there exists $\varphi$ such that $\vdash_L \varphi$ and $\not\vdash_L \neg\neg\varphi$ then (D1) does not hold.

**Proposition 6:** if there exists $\varphi$ such that $\vdash_L \neg\neg\varphi$ and $\not\vdash_L \varphi$ then (D2) does not hold.

Next, we investigate a few issues related to the condition under which a function from $\mathcal{L}$ to $[0,1]$ can be both a necessity and a possibility function.

**Definition:** a truth-functional valuation is a function $f$ from $\mathcal{L}$ to $[0,1]$ such that there exist two non-decreasing operators $\oplus$ and $\otimes$ from $[0,1]^2$ to $[0,1]$ such that $\forall \varphi, \psi$, $f(\varphi \vee \psi) = f(\varphi) \oplus f(\psi)$ and $f(\varphi \wedge \psi) = f(\varphi) \otimes f(\psi)$.

**Definition:** a logic $L$ is said to admit trivialisation of truth-functional valuations iff any truth-functional valuation $f$ satisfying (Dom), i.e. $\vdash_L \varphi \to \psi$ implies $f(\varphi) \leq f(\psi)$ (we will also say that $f$ is monotonic w.r.t. $\vdash_L$) is a classical valuation, i.e. there are two values $0^*$ and $1^*$ such that $\forall \varphi, f(\varphi) \in \{0^*, 1^*\}$ and $f(\neg\varphi) \neq f(\varphi)$.

It is well-known that trivialisation of truth-functional valuations holds in the case of classical propositional logic ([Wes 87] [DP 88] - see also [DP 94] for a discussion on the implications of this result). To study the condition under which this property also holds in the case of non-classical logics, let us consider the following assumptions:

1. $\vdash_L \varphi \to \varphi \vee \psi$
2. $\vdash_L \varphi \wedge \psi \to \varphi$
3. $\vdash_L \varphi \wedge \varphi \leftrightarrow \varphi$
4. $\vdash_L \varphi \vee \varphi \leftrightarrow \varphi$
5. $\vdash_L \varphi \wedge \psi \leftrightarrow \psi \wedge \varphi$
6. $\vdash_L \varphi \vee \psi \leftrightarrow \psi \vee \varphi$
7. $\vdash_L (\varphi \wedge \psi) \wedge \xi \leftrightarrow \varphi \wedge (\psi \wedge \xi)$
8. $\vdash_L (\varphi \vee \psi) \vee \xi \leftrightarrow \varphi \vee (\psi \vee \xi)$
9. $\vdash_L \varphi \vee \neg\varphi$ (excluded middle)
10. $\vdash_L \neg(\varphi \wedge \neg\varphi)$ (non-contradiction)

Again, an example of a logic satisfying these properties is the logic E [AB 75]. On the other hand, intuitionistic logic and paraconsistent logics do not.

**Proposition 7:** let $L$ be a logic satisfying (1) to (8) and $f$ a truth-functional valuation monotonic w.r.t. $\vdash_L$. Then we have $\oplus = \max$ and $\otimes = \min$.

**Proposition 8:** let $L$ be a logic satisfying (1) to (8) and excluded middle, and $f$ a truth-functional valuation monotonic w.r.t. $\vdash_L$. Then, $\forall \varphi, f(\varphi) = 1^*$ or $f(\neg\varphi) = 1^*$, where $1^* = \sup\{f(\varphi), \varphi \in \mathcal{L}\}$.

**Proposition 9:** let $L$ be a logic satisfying (1)-(8) and non-contradiction and $f$ a truth-functional valuation monotonic w.r.t. $\vdash_L$. Then $\forall \varphi, f(\varphi) = 0^*$ or $f(\neg\varphi) = 0^*$ where $0^* = \inf\{f(\varphi), \varphi \in \mathcal{L}\}$.

**Corollary 10:** any logic satisfying (1) to (8), excluded middle and non-contradiction admits trivialisation of truth-functional valuations.

## 2.3 Semantics of L-necessity/possibility functions

With the assumption that $\mathcal{L}$ is built from a finite number of propositional variables, "classical" necessity/possibility functions can be semantically defined by means of possibility distributions: a possibility distribution $\pi$ is simply a function from the set $\Omega$ of all interpretations for $L$ to $[0,1]$. The necessity function induced by $\pi$ is defined by $N(\varphi) = \inf\{1 - \pi(\omega) | \omega \models \neg\varphi\}$ (with the convention $\inf \emptyset = 1$ that we take in all the paper as well as $\sup \emptyset = 0$). It can then be proved that $N$ is a necessity measure, and that any necessity measure is induced by a possibility distribution.

We now turn to the general case of a logic $L$ for which the class of L-models is written $\Omega_L$.

**Definition:** a L-possibility distribution is a mapping $\pi$ from $\Omega_L$ to $[0,1]$. It is said to be normalized iff $\sup_{v \in \Omega_L} \pi(v) = 1$.

In classical logic, due to the equivalence between $v \not\models \varphi$ and $v \models \neg\varphi$, the two following definitions for inducing a C-necessity function from a C-possibility distribution are equivalent:

$$N(\varphi) = 1 - \sup\{\pi(v) | v \not\models \varphi\}$$
$$N(\varphi) = 1 - \sup\{\pi(v) | v \models \neg\varphi\}$$

Analogously, for possibility functions:

$$\Pi(\varphi) = \sup\{\pi(v) | v \not\models \varphi\}$$
$$\Pi(\varphi) = \sup\{\pi(v) | v \models \neg\varphi\}$$

**Definition:** $f_1(\pi), f_2(\pi), f_3(\pi), f_4(\pi)$ are the mappings from $\mathcal{L}$ to $[0,1]$ induced from $\pi$ by:

$$f_1(\pi)(\varphi) = 1 - \sup\{\pi(v) | v \not\models_L \varphi\}$$
$$f_2(\pi)(\varphi) = 1 - \sup\{\pi(v) | v \models_L \neg\varphi\}$$
$$f_3(\pi)(\varphi) = \sup\{\pi(v) | v \models_L \varphi\}$$
$$f_4(\pi)(\varphi) = \sup\{\pi(v) | v \not\models_L \neg\varphi\}$$

It is straightforward from these definitions that the following duality properties hold:

- $f_4(\pi)(\varphi) = 1 - f_1(\pi)(\neg\varphi)$
- $f_2(\pi)(\varphi) = 1 - f_3(\pi)(\neg\varphi)$

**Proposition 11:** if $L$ is such that $v \not\models_L \varphi \Rightarrow v \models_L \neg\varphi$ (or equivalently, $v \not\models_L \neg\varphi \Rightarrow v \models_L \varphi$)[3] then $f_2(\pi)(\varphi) \leq f_1(\pi)(\varphi)$, and $f_4(\pi)(\varphi) \leq f_3(\pi)(\varphi)$.

**Proposition 12:** $f_1$ is a L-necessity function, provided that the following conditions hold:

- if $\vdash_L \varphi$ then $v \models_L \varphi$ for all $v$ (Soundness)
- $v \models_L \varphi \leftrightarrow \psi$ iff $(v \models_L \varphi) \Leftrightarrow (v \models \psi)$
- $v \models_L \varphi \wedge \psi$ iff $v \models_L \varphi$ and $v \models_L \psi$

---

[3] Either $(v \not\models_L \varphi \Rightarrow v \models_L \neg\varphi)$ or $(v \not\models_L \neg\varphi \Rightarrow v \models_L \varphi)$ basically amounts to the validity of $\varphi \vee \neg\varphi$ in the logic $L$.

**Proposition 13:** $f_2$ is a L-necessity function, provided that the following conditions hold:

- if $\vdash_L \varphi$ then $v \not\models_L \neg\varphi$ for all $v$
- $v \models_L \varphi \leftrightarrow \psi$ iff $(v \models_L \neg\varphi) \Leftrightarrow (v \models_L \neg\psi)$
- $v \models_L \neg(\varphi \wedge \psi)$ iff $v \models_L \neg\varphi$ or $v \models_L \neg\psi$

**Proposition 14:** $f_3$ is a L-possibility function, provided that the following conditions hold:

- if $\vdash_L \varphi$ then $v \not\models_L \neg\varphi$ for all $v$
- $v \models_L \varphi \leftrightarrow \psi$ iff $(v \models_L \varphi) \Leftrightarrow (v \models_L \psi)$
- $v \models_L \varphi \vee \psi$ iff $v \models_L \varphi$ or $v \models_L \psi$

**Proposition 15:** $f_4$ is a L-possibility function, provided that the following conditions hold:

- if $\vdash_L \varphi$ then $v \models_L \varphi$ for all $v$ (Soundness)
- $v \models_L \varphi \leftrightarrow \psi$ iff $(v \models_L \neg\varphi) \Leftrightarrow (v \models_L \neg\psi)$
- $v \models_L \neg(\varphi \vee \psi)$ iff $v \models_L \neg\varphi$ and $v \models_L \neg\psi$

### 2.4 L-necessity orderings

It has been shown [Dub 86] that necessity and possibility functions can be equivalently expressed in purely qualitative terms, with preordering relations. We briefly give a generalization of this result, for the case of necessities (the case for possibilities is similar).

**Definition:** A L-necessity ordering is a relation on $\mathcal{L}$ satisfying the following properties:

- if $\varphi \geq \psi$ and $\psi \geq \xi$ then $\varphi \geq \xi$ (transitivity)
- if $\vdash_L \varphi \to \psi$ then $\psi \geq \varphi$ (dominance)
- $\varphi \wedge \psi \approx \varphi$ or $\varphi \wedge \psi \approx \psi$ (conjunctiveness)

**Definition [Dub 86]:** a relation $\geq$ on $\mathcal{L}$ is said to agree strictly with a mapping $f$ from $\mathcal{L}$ to $[0,1]$ iff $\forall \varphi, \psi \in \mathcal{L}$, we have $\varphi \geq \psi \Leftrightarrow f(\varphi) \geq f(\psi)$.

**Proposition 16** (correspondence between L-necessity functions and L-necessity orderings): the only mappings from $\mathcal{L}$ to $[0,1]$ agreeing strictly with L-necessity orderings and also satisfying (Taut) are L-necessity functions.

## 3 Application to reasoning with uncertain and inconsistent information

### 3.1 Motivations

Possibility theory, as well as its qualitative counterparts such as epistemic entrenchment relations [GM 88], ranked knowledge bases [Pea 90] or rational closure [Leh 89] provide a relativized treatment of inconsistency, since the latter becomes a gradual notion. I.e., a possibilistic knowledge base [DLP 94] consists of a set of constraints $KB = \{(\varphi_i\ \alpha_i), i = 1..n\}$, where $(\varphi_i\ \alpha_i)$ is a syntactic notation for the semantical constraint $N(\varphi_i) \geq \alpha_i$.

A possibilistic knowledge base is partially inconsistent if it leads to enforce $N(\bot) > 0$; stated otherwise, the inconsistency degree of $KB$ is defined by $Incons(KB) = \max_{S \subseteq KB, S \vdash \bot} \min_{(\varphi_i \alpha_i) \in S} \alpha_i = \min\{N(\bot), N \text{ satisfies } KB\}$. Any formula below this level, i.e. any $\varphi_i$ where $\alpha_i \leq Incons(KB)$, is then inhibited (it is "drown" by the inconsistency [BCDLP 93]). This shows that the notion of inconsistency in possibilistic logic and its qualitative counterpart is gradual but *global*. The inconsistency level measures to what extent the knowledge base is inconsistent, but do not locate the inconsistency. The aforementioned "drowning effect" is a consequence of this global treatment of inconsistency. One way to cope with it is to consider the knowledge base syntactically [Bre 89] [Neb 91] [BCDLP 93], by selecting among maximal sub-bases of $KB$ using a criterion involving the $\alpha_i$'s. However, these syntactical approaches do not have (yet) any semantics in terms of uncertainty measures.

Now, using paraconsistent logics for handling inconsistent knowledge bases enables a local treatment of inconsistency, by locating the inconsistency on some formulas. Yet, these paraconsistent approaches do not allow for any gradually in the inconsistency, which implies some loss of information if the initial knowledge was pertained with uncertainty.

While possibilistic logic allows for a gradual but global treatment of inconsistency, where conflicts are solved only by comparing the uncertainty level of the pieces of information with the inconsistency level of the knowledge base, the pure paraconsistent approach localizes inconsistency, but conflicts cannot be ranked according to uncertainty, importance, priority, normality as done in rank-based systems. Thus paraconsistency is not able to "solve" the conflicts. What we propose here is to apply the results of Section 2 to a given paraconsistent logic, namely $C_1$ [daC 74], to handle both uncertain and inconsistent knowledge, and with a local treatment of inconsistency. We now give two motivating examples, one about fusion of uncertain information (multi-source reasoning) and one about reasoning with default rules.

**Example 1** (multi-source reasoning)
This example is borrowed from [Cho 94].
Two witnesses report their observations about a murderer. Witness 1 (noted W1) is certain that the murderer was a woman with blond hair, and believes (with some uncertainty) that she was wearing a Chanel suit, glasses, and was driving a BMW. Witness 2 (noted W2) is certain that the murderer was a woman with brown hair and that she was not wearing glasses, and believes (with some uncertainty) that she was driving a Fiat.

W1 **female** (sure), **blond-hair** (sure), **drives-BMW** (unsure), **wear-glasses** (unsure), **wear-Chanel-suit** (unsure)

W2 **female** (sure), **brown-hair** (sure), **drives-Fiat** (unsure), ¬**wear-glasses** (sure)

What would we like to conclude about the following statements?

- Both witnesses agree that the murderer was female and are completely sure; so we want to conclude the murderer was female.
- No contradiction either about wear-Chanel-suit since witness 2 does not know anything.
- Strong contradiction about the colour of the murderer's hair; we wish to conclude neither blond nor brown but we want to keep in mind that these literals are "strongly subject to inconsistency" (knowing the constraint ¬(blond-hair ∧ brown-hair)).
- Contradiction about wear-glasses: the contradiction is weaker than the one above since witness 1 is unsure; moreover, since witness 2's information is prioritary to witness 1's we would like to solve the conflict (by concluding ¬wear-glasses).
- Weak contradiction again, about the car; however, since both witnesses are equally certain, we do not want to conclude anything.

**Example 2** (drowning effect)
Here, applying Pearl's ranking procedure of default rules to

$$\Delta = \{\text{penguin} \to \text{bird}, \text{penguin} \to \neg\text{fly}, \\ \text{bird} \to \text{fly}, \text{bird} \to \text{wings}\}$$

gives

$$\Delta_1 = \{\text{penguin} \to \text{bird}, \text{penguin} \to \neg\text{fly}\};$$
$$\Delta_0 = \{\text{bird} \to \text{fly}, \text{bird} \to \text{wings}\}.$$

Adding the fact penguin to $\Delta$ enables us to infer ¬fly but wings is not deduced (it is "drown" by the inconsistency appearing at rank 1). This particular case of the drowning effect is known as the property of "inheritance blocking".

Considering $\Delta$ as a set of formulas for the logic $C_1$, we obtain

$$\Delta \cup \{\text{penguin}\} \vdash_{C_1} \{\text{fly}, \neg\text{fly}, \text{wings}\}.$$

Thus, we avoid the drowning effect but we do not take into account priorities (induced by specificity) such as penguin → ¬fly over bird → fly and we conclude that fly is not well-behaved.

What we would like is to take advantage of the localisation of inconsistency, as done by paraconsistent entailment, and the priority between formulas, which would lead us to infer {fly, wings} but not ¬fly.

Note that prioritized syntax-based approaches based on the selection of maximal consistent subsets of the knowledge base guided by the priorities solve the drowning effect but do not tell anything about where the contradictions are localized; so, for instance, the conclusion ¬wear-glasses is not relativised by the fact that it is (weakly) subject to inconsistency.

### 3.2 A case study: $C_1$-necessity functions

#### 3.2.1 The paraconsistent logic $C_1$

$C_1$ [daC 74] is a paraconsistent logic, that is, a logic in which a contradiction $\varphi \wedge \neg\varphi$ fails to entail other arbitrary contradictions $\psi \wedge \neg\psi$. $C_1$ retains all inference patterns of classical logic that are not based on negation. For instance,

$$\frac{\varphi \quad \psi}{\varphi \wedge \psi}$$

is valid in $C_1$. By contrast, some inference patterns of classical logic that do appeal to negation are not preserved. For instance,

$$\frac{\neg\varphi \quad \neg\psi}{\neg(\varphi \vee \psi)}$$

is not valid in $C_1$. The idea is that positive information is fundamental: positive formulas and inferences contribute to state what the facts are whereas negative formulas and inferences are merely constraints (in the sense of integrity constraints for databases). Accordingly, $C_1$ allows us to elicit all and only the formulas responsible for a given contradiction ([CL 92] [BL 94]).

A valuation-based semantics for $C_1$ [Alv 84] is given in Section 3.2.3 as we now reproduce the original axiomatic presentation of $C_1$ that consists of the next ten axioms

1. $\varphi \to (\psi \to \varphi)$
2. $(\varphi \to \psi) \to [(\varphi \to (\psi \to \sigma)) \to (\varphi \to \sigma)]$
3. $\varphi \wedge \psi \to \varphi$
4. $\varphi \wedge \psi \to \psi$
5. $\varphi \to (\psi \to \varphi \wedge \psi)$
6. $\varphi \to \varphi \vee \psi$
7. $\varphi \to \psi \vee \varphi$
8. $(\varphi \to \sigma) \to [(\psi \to \sigma) \to (\varphi \vee \psi \to \sigma)]$
9. $\varphi \vee \neg\varphi$
10. $\neg\neg\varphi \to \varphi$

together with the single inference rule $\frac{\varphi \quad \varphi \to \psi}{\psi}$.

$C_1$ has the following basic features. First, the connectives are not interdefinable. For instance, $\varphi \vee \psi$ cannot be defined as $\neg(\neg\varphi \wedge \neg\psi)$. Second, the replacement of equivalent formulas does not hold. For instance, $(\varphi \vee \psi) \leftrightarrow [(\varphi \to \psi) \to \psi]$ is valid in $C_1$ but $\neg(\varphi \vee \psi) \leftrightarrow \neg[(\varphi \to \psi) \to \psi]$ is not. Third, neither modus tollens

$$\frac{\varphi \to \psi \quad \neg\psi}{\neg\varphi}$$

nor disjunctive syllogism

$$\frac{\varphi \vee \psi \quad \neg\varphi}{\psi}$$

are valid in $C_1$.

Regarding notation, we use $\varphi^\circ$ as an abbreviation for $\neg(\varphi \wedge \neg\varphi)$. In the next two sections, we also use # to denote any of $\wedge, \vee, \to$.

### 3.2.2  $C_1$-necessity functions: definition and basic properties

**Definition:** like for L-necessity functions, replacing $\vdash_L$ by $\vdash_{C_1}$.

Some properties enjoyed by $C_1$-necessity functions are:

Dom: If $\vdash_{C_1} \varphi \to \psi$ then $N(\varphi) \leq N(\psi)$

(P1) $N(\varphi) = N(\neg\varphi^\circ)$ or $N(\neg\varphi) = N(\neg\varphi^\circ)$

(P2) $N(\psi) \geq \min(N(\varphi), N(\varphi \to \psi))$

(P3) $\varphi_1, \ldots, \varphi_n \vdash_{C_1} \psi \Rightarrow N(\psi) \geq \min_{i=1..n} N(\varphi_i)$

(P4) $N(\neg\varphi) \geq \min(N(\varphi^\circ), N(\varphi \to \psi), N(\varphi \to \neg\psi))$

(P5) $N(\varphi \vee \psi) \geq \max(N(\varphi), N(\psi))$

(P6) $N(\neg\neg\varphi) = N(\varphi)$

(P7) $N(\varphi^{\circ\circ}) = 1$

(P8) $N((\varphi \# \psi)^\circ) \geq \min(N(\varphi^\circ), N(\psi^\circ))$

(P9) $N$ is a classical necessity function if and only if $N(\varphi^\circ) = 1$ for all $\varphi$

$N(\neg\varphi^\circ) = \min(N(\varphi), N(\neg\varphi))$ is the necessity of $\varphi$ "behaving badly"; it can be seen as a measure of the inconsistency inherent in $\varphi$. $C_1$-necessity functions enable us to rank the formulas not only with respect to their certainty, but also with respect to their inherent inconsistency: $N(\neg\varphi^\circ)$ gives a notion of inconsistency which is both local and gradual. We recover of course as particular cases:

- Classical necessity functions, so that $N(\neg\varphi^\circ) = N(\bot)$ for all $\varphi$. The notion of inconsistency is still gradual but global.
- Classical $C_1$-valuations, which verify $N(\neg\varphi^\circ) = 0$ or $N(\neg\varphi^\circ) = 1$, for all $\varphi$. The notion of inconsistency is still local but not gradual.

### 3.2.3  $C_1$-necessity functions: semantics

At first, a (paraconsistent) $C_1$-valuation [Alv 84] is a mapping from $\mathcal{L}$ to $\{0,1\}$ such that:

- $v(\varphi) = 0 \Rightarrow v(\neg\varphi) = 1$
- $v(\neg\neg\varphi) = 1 \Leftrightarrow v(\varphi) = 1$
- $v(\psi^\circ) = v(\varphi \to \psi) = v(\varphi \to \neg\psi) = 1 \Rightarrow v(\varphi) = 0$
- $v(\varphi \to \psi) = 1 \Leftrightarrow v(\varphi) = 0$ or $v(\psi) = 1$
- $v(\varphi \wedge \psi) = 1 \Leftrightarrow v(\varphi) = 1$ and $v(\psi) = 1$
- $v(\varphi \vee \psi) = 1 \Leftrightarrow v(\varphi) = 1$ or $v(\psi) = 1$
- $v(\varphi^\circ) = v(\psi^\circ) = 1 \Rightarrow v((\varphi \# \psi)^\circ) = 1$

**Definition:** a $C_1$-possibility distribution is a mapping $\pi$ from the set of all $C_1$-valuations to $[0,1]$.

Due to Proposition 12, the function $f_1(\pi)$ defined as

$$f_1(\pi)(\varphi) = 1 - \sup\{\pi(v) | v(\varphi) = 0\}$$

is a $C_1$-necessity function (since $C_1$ obeys the condition stated in Proposition 12 - the soundness of the semantics coming from the soundness and completeness of $C_1$ established in [Alv 84]).

We could have also defined $C_1$-possibility functions, $C_1$-necessity and possibility orderings, that we do not discuss for the sake of brevity. $C_1$-necessity functions are sufficient to deal with the next section, devoted to the application to reasoning with uncertain and inconsistent knowledge.

### 3.3  Reasoning with $C_1$-necessity functions

#### 3.3.1  Generalizing the principle of minimum specificity

The principle of minimum specificity [DP 86] or equivalently, minimum compact ranking [Pea 90] and rational closure [Leh 89] (all these being equivalent, up to the language on which they are defined) induces, from a possibilistic knowledge base, a particular necessity function i.e. the smallest among all necessity functions satisfying the knowledge base. Thanks to the property (P3), we are able to generalize the principle of minimum specificity to $C_1$-necessity functions:

**Definition:** a $C_1$-possibilistic knowledge base is a finite set $KB = \{(\varphi_i \; \alpha_i), 1 \leq i \leq n\}$ where $\varphi_i \in \mathcal{L}$ and $\alpha_i \in [0,1]$. A $C_1$-necessity function $N$ is said to satisfy $KB$ iff $\forall i = 1..n, N(\varphi_i) \geq \alpha_i$.

**Definition:** the minimum specificity closure $N_{KB}$ of a $C_1$-possibilistic knowledge base $KB$ is the $C_1$-necessity function defined by

$$\forall \psi \in \mathcal{L}, N_{KB}(\psi) = \sup\{\beta | KB_\beta \vdash_{C_1} \psi\}$$

where $KB_\beta = \{\varphi_i | (\varphi_i \; \alpha_i) \in KB \text{ and } \alpha_i \geq \beta\}$.

**Proposition 17** (principle of minimum specificity for $C_1$-necessities): For any $C_1$-necessity function $N$, $N$ satisfies $KB$ iff $N \geq N_{KB}$.

More generally, the minimum specificity closure could be extended to any logic $L$ satisfying the property (P3). Applying the principle of minimum specificity enables us to draw conclusions that taking into account the uncertainty and the inconsistency of the knowledge base. We propose the following definition of a consequence relation:

**Definition:** $KB \mathrel{|\!\!\sim} \psi$ iff $N_{KB}(\psi) > N_{KB}(\neg\psi^\circ)$.

**Proposition 18:** $KB \mathrel{|\!\!\sim} \psi$ iff $N_{KB}(\psi) > N_{KB}(\neg\psi)$.

Intuitively, we deduce $\psi$ from $KB$ iff the certainty of $\psi$ is higher than the inconsistency inherent to $\psi$, or equivalently, iff the certainty of $\psi$ is higher than the certainty of $\neg\psi$. The binary version of $\mathrel{|\!\!\sim}$ would be defined by $\varphi \mathrel{|\!\!\sim}_{KB} \psi$ iff $N_{KB}(\varphi \to \psi) > N_{KB}(\varphi \to \neg\psi^\circ)$, or equivalently iff $N_{KB}(\varphi \to \psi) > N_{KB}(\varphi \to \neg\psi)$. Note that $\mathrel{|\!\!\sim}$ is nonmonotonic; a more complete study of the properties of $\mathrel{|\!\!\sim}$ à la Kraus, Lehmann and Magidor [KLM 90], is possible with respect to the (monotonic) logic $C_1$ instead of classical logic.

Note that when $N$ collapses to a classical necessity measure, we have $\forall \psi N_{KB}(\neg\psi^\circ) = N(\bot)$ and $\mathrel{|\!\!\sim}$ is the classical possibilistic consequence relation [DP 91b].

**Example** (multi-source reasoning): let us return to the example of Section 3.1. Taking some $\alpha \in (0,1)$:

- W1 (witness 1):
  $N(\text{female}) = 1$; $N(\text{brown}) = 1$; $N(\text{BMW}) = \alpha$; $N(\text{Chanel}) = \alpha$; $N(\text{glasses}) = \alpha$.
- W2 (witness 2):
  $N(\text{female}) = 1$; $N(\neg\text{brown}) = 1$; $N(\neg\text{BMW}) = \alpha$; $N(\neg\text{glasses}) = 1$.

The fusion $KB$ of these two knowledge bases gives the following minimum specificity closure:

- $N_{KB}(\text{female}) = 1$; $N_{KB}(\neg\text{female}) = 0$; $N_{KB}(\neg\text{female}^o) = 0$;
- $N_{KB}(\text{brown}) = 1$; $N_{KB}(\neg\text{brown}) = 1$; $N_{KB}(\neg\text{brown}^o) = 1$;
- $N_{KB}(\text{BMW}) = \alpha$; $N_{KB}(\neg\text{BMW}) = \alpha$;
- $N_{KB}(\text{Chanel}) = \alpha$; $N_{KB}(\neg\text{Chanel}) = 0$; $N_{KB}(\neg\text{Chanel}^o) = 0$;
- $N_{KB}(\text{glasses}) = \alpha$; $N_{KB}(\neg\text{glasses}) = 1$; $N_{KB}(\neg\text{glasses}^o) = \alpha$.

Therefore, we have $KB \hspace{1pt}\vdash\hspace{-6pt}\sim\hspace{2pt} \text{female}$, $KB \hspace{1pt}\vdash\hspace{-6pt}\sim\hspace{2pt} \text{Chanel}$, $KB \hspace{1pt}\vdash\hspace{-6pt}\sim\hspace{2pt} \neg\text{glasses}$; however, $KB \hspace{1pt}\not\vdash\hspace{-6pt}\sim\hspace{2pt} \text{BMW}$, $KB \hspace{1pt}\not\vdash\hspace{-6pt}\sim\hspace{2pt} \neg\text{BMW}$, $KB \hspace{1pt}\not\vdash\hspace{-6pt}\sim\hspace{2pt} \text{brown}$, $KB \hspace{1pt}\not\vdash\hspace{-6pt}\sim\hspace{2pt} \neg\text{brown}$.

### 3.3.2 Handling default rules

**Example:** Consider the fact penguin and the rules

$\Delta = \{\text{penguin} \to \text{bird}, \text{penguin} \to \neg\text{fly},$
$\text{bird} \to \text{fly}, \text{bird} \to \text{wings},$
$\text{fly} \to \neg\text{live-in-Antarctica}\}.$

Applying the Z ranking procedure to $\Delta$ (written with the possibilistic ranking convention) gives the ranking: (for any $\alpha, \beta$ such that $0 < \beta < \alpha < 1$)

$\Delta_\alpha = \{\text{penguin} \to \text{bird}, \text{penguin} \to \neg\text{fly},$
$\text{fly} \to \neg\text{live-in-Antarctica}\};$
$\Delta_\beta = \{\text{bird} \to \text{fly}, \text{bird} \to \text{wings}\}.$

Then, taking the $C_1$-minimum specificity closure of $KB = \{\text{penguin}\} \cup \Delta$ leads to

- $N_{KB}(\text{penguin}) = 1$;
- $N_{KB}(\text{bird}) = \alpha$;
  $N_{KB}(\neg\text{fly}) = \alpha$;
  $N_{KB}(\text{fly} \to \neg\text{live-in-Antarctica}) = \alpha$;
- $N_{KB}(\text{fly}) = \beta$;
  $N_{KB}(\neg\text{fly}^o) = \beta$;
  $N_{KB}(\text{wings}) = \beta$;
  $N_{KB}(\neg\text{live-in-Antarctica}) = \beta$;
- $N_{KB}(\neg\text{bird}) = 0$;
  $N_{KB}(\neg\text{bird}^o) = 0$;
  $N_{KB}(\neg\text{penguin}) = 0$;
  $N_{KB}(\neg\text{penguin}^o) = 0$;
  $N_{KB}(\neg\text{wings}) = 0$;
  $N_{KB}(\neg\text{wings}^o) = 0$;
  $N_{KB}(\text{live-in-Antarctica}) = 0$;
  $N_{KB}(\neg\text{live-in-Antarctica}^o) = 0$.

Therefore, we have

$KB \hspace{1pt}\vdash\hspace{-6pt}\sim\hspace{2pt} \neg\text{fly}$ (which is intended);

$KB \hspace{1pt}\vdash\hspace{-6pt}\sim\hspace{2pt} \neg\text{wings}$ (which is intended);
$\hspace{1pt}\vdash\hspace{-6pt}\sim\hspace{2pt}$ avoids the drowning effect, contrarily to the classical minimum specificity closure, System Z, and similar systems.

but also

$KB \hspace{1pt}\vdash\hspace{-6pt}\sim\hspace{2pt} \neg\text{live-in-Antarctica}$

which is not intended! (Due to $N_{KB}(\text{fly}) = \beta$, the rule $\text{fly} \to \neg\text{live-in-Antarctica}$ applies).

Here is a revised definition, more suited to handling default rules:

**Definition:** Let $KB = F \cup \Delta$, where $F$ is a set of facts and $\Delta = \{\varphi_i \to \psi_i, i = 1..n\}$ a set of default rules, where each rule is assigned a necessity degree corresponding to its Z-ranking. We define

$G^0(\Delta) = F \cup \Delta$

and $\forall k \geq 0$,

$G^{k+1}(\Delta) =$
$F \cup \{\varphi_i \to \psi_i \in G^k(\Delta) \mid N_{G^k(\Delta)}(\varphi_i) > N_{G^k(\Delta)}(\neg\varphi_i^o)\}$
$= F \cup \{\varphi_i \to \psi_i \in G^k(\Delta) \mid G^k(\Delta) \hspace{1pt}\vdash\hspace{-6pt}\sim\hspace{2pt} \varphi_i\}$.

Lastly, let $G^\infty(\Delta) = \cap_{k \geq 0} G^k(\Delta)$. Then $\Delta \hspace{1pt}\vdash\hspace{-6pt}\sim\hspace{2pt}^\bullet \psi$ iff $G^\infty(\Delta) \hspace{1pt}\vdash\hspace{-6pt}\sim\hspace{2pt} \psi$.

**Example:** We apply the usual ranking procedure:

$\Delta = \{\text{bird} \to \text{fly}, \text{bird} \to \text{wings},$
$\text{penguin} \to \text{bird}, \text{penguin} \to \neg\text{fly},$
$\text{fly} \to \neg\text{live-in-Antarctica}\}.$

$G^1(\Delta) = \{\text{penguin}, \text{bird} \to \text{fly}, \text{bird} \to \text{wings},$
$\text{penguin} \to \text{bird}, \text{penguin} \to \neg\text{fly}\}.$

Clearly, $G^\infty(\Delta) = G^1(\Delta)$. Therefore, $\Delta \hspace{1pt}\vdash\hspace{-6pt}\sim\hspace{2pt}^\bullet \neg\text{fly}$. Also, $\Delta \hspace{1pt}\vdash\hspace{-6pt}\sim\hspace{2pt}^\bullet \text{bird}$ and $\Delta \hspace{1pt}\vdash\hspace{-6pt}\sim\hspace{2pt}^\bullet \text{wings}$. Contrastedly, $\Delta \hspace{1pt}\not\vdash\hspace{-6pt}\sim\hspace{2pt}^\bullet \neg\text{live-in-Antarctica}$.

## 4 Conclusion

We have given some basic results describing what remains and what changes when switching from classical possibility theory to possibility theory over a non-classical logic. We have then focused on a case study, namely the paraconsistent logic $C_1$, and showed how to use it to reason with inconsistent and uncertain information. What has been left aside in this paper is the other possible applications of possibility theory over non-classical logics: first, one could think of applying the general results of Section 2 to other non-classical logics: for instance, introducing possibility and necessity valuations into intuitionistic logic could model gradual strengths of proofs; or, introducing them to Kleene's logic (or more generally to a multi-valued logic) would enable us to handle both uncertainty and partial truth.

Another topic for further research would be a parallel study for other numerical theories of uncertainty. For instance, paraconsistent probabilities would lead to a more quantitative framework for reasoning with uncertain and conflictual information; in this framework, noticing that $Prob(\varphi) + Prob(\neg\varphi) = Prob(\varphi \vee \neg\varphi) + Prob(\varphi \wedge \neg\varphi) = 1 + Prob(\neg\varphi^\circ)$, relaxing the constraint $Prob(\neg\varphi^\circ) = 0$ would make $Prob(\varphi) + Prob(\neg\varphi) > 1$ possible for some formulas; then one could think of searching for the "least inconsistent" probability distribution satisfying a set of constraints, which could be useful for instance when rectifying a set of inconsistent probabilistic data.

## 5   References

[Alv 84] E. H. Alves, "Paraconsistent Logic and Model Theory", *Studia Logica* 43, 17-32, 1984.

[AB 75] A. R. Anderson, N. D. Belnap Jr., *Entailment: The Logic of Relevance and Necessity*, Vol. 1, Princeton University Press, 1975.

[BCDLP 93] S. Benferhat, C. Cayrol, D. Dubois, J. Lang, H. Prade, "Inconsistency Management and Prioritized Syntax-Based Entailment", Proc. of the 13th Int. Joint Conf. on Artificial Intelligence (IJCAI-93), Chambéry (France), 640-645, 1993.

[BDP 92] S. Benferhat, D. Dubois, H. Prade, "Representing Default Rules in Possibilistic Logic", Proc. of the 3rd Conf. on Principles of Knowledge Representation and Reasoning (KR'92), 673-684, 1992.

[BL 94] Ph. Besnard, E. Laenens, "A Knowledge Representation Perspective: Logics for Paraconsistent Reasoning", *International Journal of Intelligent Systems* 9, 153-168, 1994.

[Bou 92] C. Boutilier, "Modal Logics for Qualitative Possibility and Beliefs", Proc. of Uncertainty in Artificial Intelligence (UAI-92), 17-23, 1992.

[Bre 89] G. Brewka, "Preferred Subtheories: An Extended Logical Framework for Default Reasoning", Proc. of the 11th Int. Joint Conf. on Artificial Intelligence (IJCAI-89), Detroit, 1043-1048, 1989.

[CL 92] W. A. Carnielli, M. Lima-Marques, "Reasoning under Inconsistent Knowledge", *Applied Non-Classical Logics* 2, 49-79, 1992.

[Cho 94] L. Cholvy, "Fusion de sources d'information ordonnées en fonction des thèmes", Proc. of RFIA'94 (in French), Paris, 487-494, 1994.

[daC 74] N. C. A. da Costa, "On the Theory of Inconsistent Formal Systems", *Notre Dame Journal of Formal Logic* 15, 497-510, 1974.

[Dub 86] D. Dubois, "Belief Structures, Possibility Theory, Decomposable Confidence Measures on Finite Sets", *Computers and Artificial Intelligence* 5, 403-417, 1986.

[DP 86] D. Dubois, H. Prade, "The Principle of Minimum Specificity as a Basis for Evidential Reasoning", in *Uncertainty in Knowledge Bases* (B. Bouchon-Meunier, R. Yager, Eds.), 75-84, 1986.

[DP 88] D. Dubois, H. Prade, "An Introduction to Possibilistic and Fuzzy Logics (with discussions). In *Non-Standard Logics for Automated Reasoning* (P. Smets et al., eds.), 287-315. Academic Press, 1988.

[DP 91a] D. Dubois, H. Prade, "Epistemic Entrenchment and Possibilistic Logic", *Artificial Intelligence* 50, 223-239, 1991.

[DP 91b] D. Dubois, H. Prade, "Possibilistic Logic, Preferential Models, Nonmonotonicity and Related Issues", Proc. of the 12th Int. Joint Conf. on Artificial Intelligence (IJCAI-91), Sydney, 419-424, 1991.

[DP 94] D. Dubois, H. Prade, "Can we Enforce Full Compositionality in Uncertainty Calculi?", Proc. of AAAI-94, to appear.

[DLP 94] D. Dubois, J. Lang, H. Prade, "Possibilistic Logic", in: *Handbook of Logic in Artificial Intelligence and Logic Programming* (D. M. Gabbay, C. Hogger, J. A. Robinson, Eds.), Oxford University Press, 439-513.

[FHL 94] L. Fariñas del Cerro, A. Herzig, J. Lang, "From Ordering-Based Nonmonotonic Reasoning to Conditional Logics", *Artificial Intelligence* 66, 375-393, 1994.

[GM 88] P. Gärdenfors, D. Makinson, "Revision of Knowledge Systems using Epistemic Entrenchment", Proc. of the 2nd Conf. on Theoretical Aspects of Reasoning about Knowledge (TARK-88), 83-95, 1988.

[KLM 90] S. Kraus, D. Lehmann, M. Magidor, "Nonmonotonic Reasoning, Preferential Models and Cumulative Logics", *Artificial Intelligence* 44, 167-207, 1990.

[Leh 89] D. Lehmann, "What does a Conditional Knowledge Base Entail?", Proc. of the 1st Conf. on Principles of Knowledge Representation and Reasoning (KR-89), Toronto, 357-367, 1989.

[Neb 91] B. Nebel, "Belief Revision and Default Reasoning: Syntax-Based Approaches", Proc. of the 2nd Conf. on Principles of Knowledge Representation and Reasoning (KR-91), Cambridge (MA), 417-428, 1991.

[Pea 90] J. Pearl, "System Z: A Natural Ordering of Defaults with Tractable Applications to Nonmonotonic Reasoning", Proc. of the 3rd Conf. on Theoretical Aspects of Reasoning about Knowledge (TARK-90), 121-135, 1990.

[Wes 87] T. Weston, "Approximate Truth", *Journal of Philosophical Logic* 16, 203-227, 1987.

[Zad 78] L. A. Zadeh, "Fuzzy Sets as a Basis for a Theory of Possibility", *Fuzzy Sets and Systems* 1, 3-28, 1978.